\documentclass[11pt]{article}
\usepackage{acl2015}
\usepackage{times}
\usepackage{url}
\usepackage{latexsym}
\usepackage{graphicx}

\title{Reward Shaping with Recurrent Neural Networks for Speeding up On-Line Policy Learning in Spoken Dialogue Systems}

\author{Pei-Hao Su, David Vandyke,  Milica Ga{\v{s}}i\'c,  \\ {\bf Nikola Mrk{\v{s}}i\'c, Tsung-Hsien Wen \and  Steve Young} \\
  Department of Engineering, University of Cambridge, Cambridge, UK\\
  {\tt \{phs26, djv27, mg436, nm480, thw28, sjy\}@cam.ac.uk}\
 }

\date{}

\begin{document}
\maketitle
\begin{abstract}

Statistical spoken dialogue systems 
have the attractive property of being able to be optimised from data via interactions with real users. However in the reinforcement learning paradigm the dialogue manager (agent) often requires significant time to explore the state-action space to learn to behave in a desirable manner. This is a critical issue when the system is trained on-line with real users where learning costs are expensive.
Reward shaping is one promising technique for addressing these concerns. Here we examine three recurrent neural network (RNN) approaches for providing reward shaping information in addition to the primary (task-orientated) environmental feedback. These RNNs are trained on returns from dialogues generated by a simulated user and attempt to diffuse the overall evaluation of the dialogue back down to the turn level to guide the agent towards good behaviour faster. In both simulated and real user scenarios these RNNs are shown to increase policy learning speed.  Importantly, they do not require prior knowledge of the user's goal. 


\end{abstract}

\section{Introduction}
\label{sec:intro}
Spoken dialogue systems (SDS) offer a natural way for people to interact with computers. With the ability to learn from data (interactions) statistical SDS can theoretically be created faster and with less man-hours than a comparable handcrafted rule based system. They have also been shown to perform better \cite{POMDP-review}. Central to this is the use of partially observable Markov decision processes (POMDP) to model dialogue, which inherently manage the uncertainty created by errors in speech recognition and semantic decoding \cite{POMDP_williams}.

The dialogue manager is a core component of an SDS and largely determines the quality of interaction. Its behaviour is controlled by a \textit{policy} which maps belief states to system actions (or distributions over sets of actions) and this policy is trained in a reinforcement learning framework \cite{RL} where rewards are received from the environment, the most informative of which occurs only at the dialogues conclusion, indicating task success or failure.\footnotemark{}\footnotetext{A uniform reward of -1 is common for all other, non-terminal turns, which promotes faster task completion.}


It is the sparseness of this environmental reward function which, by not providing any information at intermediate turns, requires exploration to traverse deeply many sub-optimal paths. This is a significant concern when training SDS on-line with real users where one wishes to minimise client exposure to sub-optimal system behaviour. In an effort to counter this problem, \textit{reward shaping} \cite{RS_Andrew} introduces domain knowledge to provide earlier informative feedback to the agent (additional to the environmental feedback) for the purpose of biasing exploration for discovering optimal behaviour quicker.\footnotemark{} Reward shaping is briefly reviewed in Section \ref{sub:rs}.

In the context of SDS, \newcite{FerreiraL15} have motivated the use of reward shaping via analogy to the `social signals' naturally produced and interpreted throughout a human-human dialogue. 
This non-statistical reward shaping model used heuristic features for speeding up policy learning.


\footnotetext{Learning algorithms are another central element in improving the speed of convergence during policy training. In particular the sample-efficiency of the learning algorithm can be the deciding factor in whether it can realistically be employed on-line. See e.g. the GP-SARSA \cite{GPRL} and Kalman temporal-difference \cite{KTD} methods which bootstrap estimates of sparse value functions from minimal numbers of samples (dialogues).}

As an alternative, one may consider attempting to handcraft a finer grained environmental reward function. For example, \newcite{TCTL} proposed diffusing expert ratings of dialogues to the state transition level to produce a richer reward function. Policy convergence may occur faster in this altered POMDP and dialogues generated by a task based simulated user may also alleviate the need for expert ratings. However, unlike reward shaping, modifying the environmental reward function also modifies the resulting optimal policy.


We recently proposed convolutional and recurrent neural network (RNN) approaches for determining dialogue success. This was used to provide a reinforcement signal for learning on-line from real users without requiring any prior knowledge of the user's task \cite{Su_2015}. Here we extend the RNN approach by introducing new training constraints in order to combine the merits of the above three works: (1) diffusing dialogue level ratings down to the turn level to  (2) add reward shaping information for faster policy learning, whilst (3) not requiring prior task knowledge which is simply unavailable on-line. 


In Section \ref{sec:models} we briefly describe potential based reward shaping before introducing the RNNs we explore for producing reward shaping signals (basic RNN, long short-term memory (LSTM) and gated recurrent unit (GRU)). The features the RNNs use along with the training constraint and loss are also described. The experimental evaluation is then presented in Section \ref{sec:exp}. Firstly, the estimation accuracy of the RNNs is assessed. The benefit of using the RNN for reward shaping in both simulated and real user scenarios is then also demonstrated. Finally, conclusions are presented in Section \ref{sec:conclusions}.

\section{RNNs for Reward Shaping}
\label{sec:models}

\subsection{Reward Shaping}
\label{sub:rs}

Reward shaping provides the system with an extra reward signal $F$ in addition to environmental reward $R$, making the system learn from the composite signal $R + F$. The shaping reward $F$ often encodes expert knowledge that complements the sparse signal $R$. Since the reward function defines the system's objective, changing it may result in a different task. When the task is modelled as a fully observable Markov decision process (MDP), \newcite{RS_Andrew} defined formal requirements on the shaping reward as a difference of any potential function $\phi$ on consecutive states $s$ and $s'$ which preserves the optimality of policies. Based on this property, \newcite{POMDP_RS} further extended it to POMDP by proof and empirical experiments:

\vspace{-3mm}
\begin{equation}
F(b_t, a, b_{t+1}) = \gamma \phi(b_{t+1}) - \phi(b_{t})
\label{eqn:reward_shaping}
\end{equation}
where $\gamma$ is the discount factor, $b_t$ the belief state at turn $t$, and $a$ the action leading $b_t$ to $b_{t+1}$.


However determining an appropriate potential function for an SDS is non-trivial. Rather than hand-crafting the function with heuristic knowledge, we propose using an RNN to predict proper values as in the following.

\subsection{Recurrent Neural Network Models}
\label{sec:rnns}

\begin{figure}[t]
\centerline{\includegraphics[width=75mm]{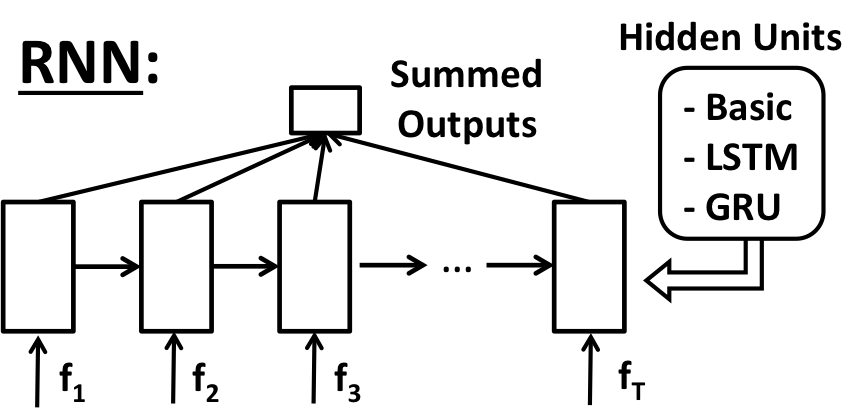}}
\caption{{RNN with three types of hidden units: basic, LSTM and GRU. The feature vectors ${\bf f}_t$ extracted at turns $t=1,\dots,T$ are labelled ${\bf f}_t$.}} 
\vspace{-3mm}
\label{fig:models}
\end{figure}

The RNN model is a subclass of neural network defined by the presence of feedback connections. The ability to succinctly retain history information makes it suitable for modelling sequential data. It has been successfully used for language modelling  \cite{mikolov2011extensions} and spoken language understanding \cite{RNNonSLU}.

However, \newcite{bengio1994learning} observed that basic RNNs suffer from vanishing/exploding gradient problems that limit their capability of modelling long context dependencies. To address this, long short-term memory \cite{hochreiter1997long} and gated recurrent unit \cite{chung2014empirical} RNNs have been proposed. In this paper, all three types of RNN (basic/LSTM/GRU) are compared.


\subsection{Reward Shaping with RNN Prediction}
\label{sub:RNNRS}

The role of the RNN is to solve the regression problem of predicting the scalar return of each completed dialogue. At every turn $t$, input feature $f_t$ are extracted from the belief/action pair and used to update the hidden layer $h_t$. From dialogues generated by a simulated user \cite{userSim} supervised training pairs are created which consist of the turn level sequence of these feature vectors $f_t$ along with the scalar dialogue return as scored by an objective measure of task completion. Whilst the RNN models are trained on dialogue level supervised targets, we hypothesise that their subsequent turn level predictions can guide policy exploration via acting as informative reward shaping potentials. 

To encourage good turn level predictions, all three RNN variants are trained to predict the dialogue return not with the final output of the network, but with the constraint that their scalar outputs from each turn $t$ should sum to predict the return for the whole dialogue. This is shown in Figure \ref{fig:models}. A mean-square-error (MSE) loss is used (see Appendix \ref{appendix}). The trained RNNs are then used directly as the reward shaping potential function $\phi$, using the RNN scalar output at each turn. 



The feature inputs $f_t$ for all RNNs consisted of the following sections: the real-valued belief state vector formed by concatenating the distributions over user discourse act, method and goal variables \cite{BUDS}, one-hot encodings of the user and summary system actions, and the normalised turn number. This feature vector was extracted at every turn (system + user exchange).

\section{Experiments}
\label{sec:exp}

\subsection{Experimental Setup}

In all experiments the Cambridge restaurant domain was used, which consists of approximately 150 venues each having 6 attributes (slots) of which 3 can be used by the system to constrain the search and the remaining 3 are informable properties once a database entity has been found. 

The shared core components of the SDS in all experiments were a domain independent ASR, a confusion network (CNet) semantic input decoder \cite{CNET}, the BUDS \cite{BUDS} belief state tracker that factorises the dialogue state using a dynamic Bayesian network and a template based natural language generator. All policies were trained by GP-SARSA \cite{GPRL} and the summary action space contains 20 actions. Per turn reward was set to -1 and final reward 20 for task success else 0.

With this ontology, the size of the full feature vector was 147. The turn number was expressed as a percentage of the maximum number of allowed turns, here 30. The one-hot user dialogue act encoding was formed by taking only the most likely user act estimated by the CNet decoder.

\subsection{Neural Network Training}
\label{sub:nnt}

\begin{figure}[t]
\centerline{\includegraphics[width=80mm]{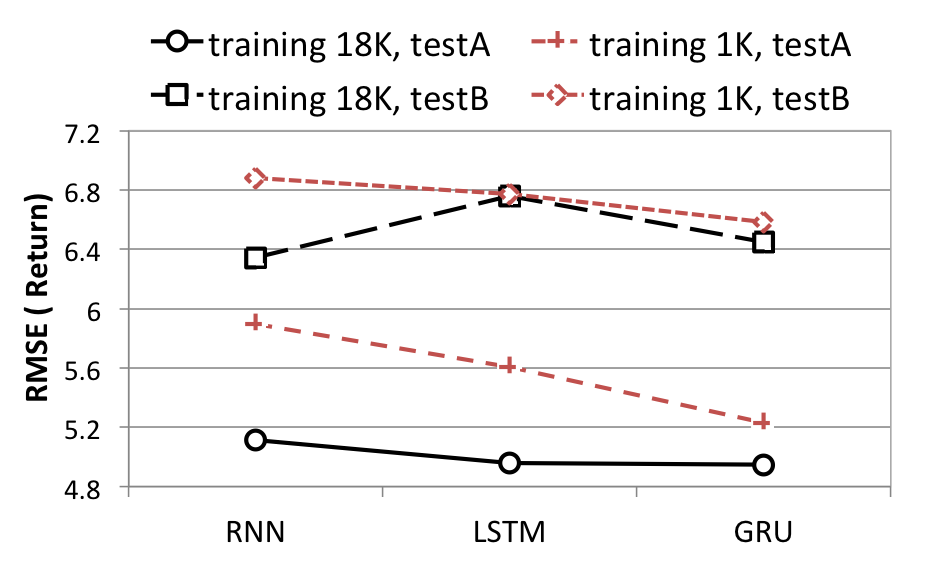}}
\vspace{-3mm}
\caption{{RMSE of return prediction by using RNN/LSTM/GRU, trained on 18K and 1K dialogues and tested on sets \textit{testA} and \textit{testB} (see text).}}
\vspace{-3mm}
\label{fig:NN_training}
\end{figure}

Here results of training the 3 RNNs on the simulated user dialogues are presented.\footnotemark{} Two training sets were used consisting of 18K and 1K dialogues to verify the model robustness. In all cases a separate validation set consisting of 1K dialogues was used for controlling overfitting. Training and validation sets were approximately balanced regarding objective success/failure labels and collected at a 15\% semantic error rate (SER). 
Prediction results are shown in Figure \ref{fig:NN_training} on two test sets; \textit{testA:} 1K dialogues, balanced regarding objective labels, at 15\% SER and \textit{testB:} containing 12K dialogues collected at SERs of $0,15,30$ and $45$ as the data occurred (\textit{i.e.} with no balancing regarding labels). 

\footnotetext{All RNNs were implemented using the Theano library \cite{bergstra+al:2010-scipy}. In all cases the hidden layer contained 100 units with a sigmoid non-linearity and used stochastic gradient descent (per dialogue) for training.}

Root-MSE (RMSE) results of predicting the dialogue return are depicted in Figure \ref{fig:NN_training}. The models with LSTM and GRU units achieved a slight improvement in most cases over the basic RNN. Notice that the model with GRU even reached comparable results when trained with 1K training data compared to 18K.
The results from the 1K training set indicate that the model can be developed from limited data. This enables datasets to be created by human annotation, avoiding the need for a simulated user.
The results on set \textit{testB} also show that the models can perform well in situations with varying error rates as would be encountered in real operating environments. 
Note that the dataset could also be created from human's annotation which avoids the need for a simulated user.
We next examine the RNN-based reward shaping for policy training with a simulated user.


\subsection{Policy Learning with Simulated User}
\label{sub:sim}

Since the aim of reward shaping is to enhance policy learning speed, we focus on the first 1000 training dialogues. Figure \ref{fig:NN_training} shows that the GRU RNN attained slightly better performance than the other two RNN models, albeit with no statistical significance. Thus for clearer presentation of the policy training results we plot only the GRU results, using the model trained on 18K dialogues.  

\begin{figure}[t]
\centerline{\includegraphics[width=80mm]{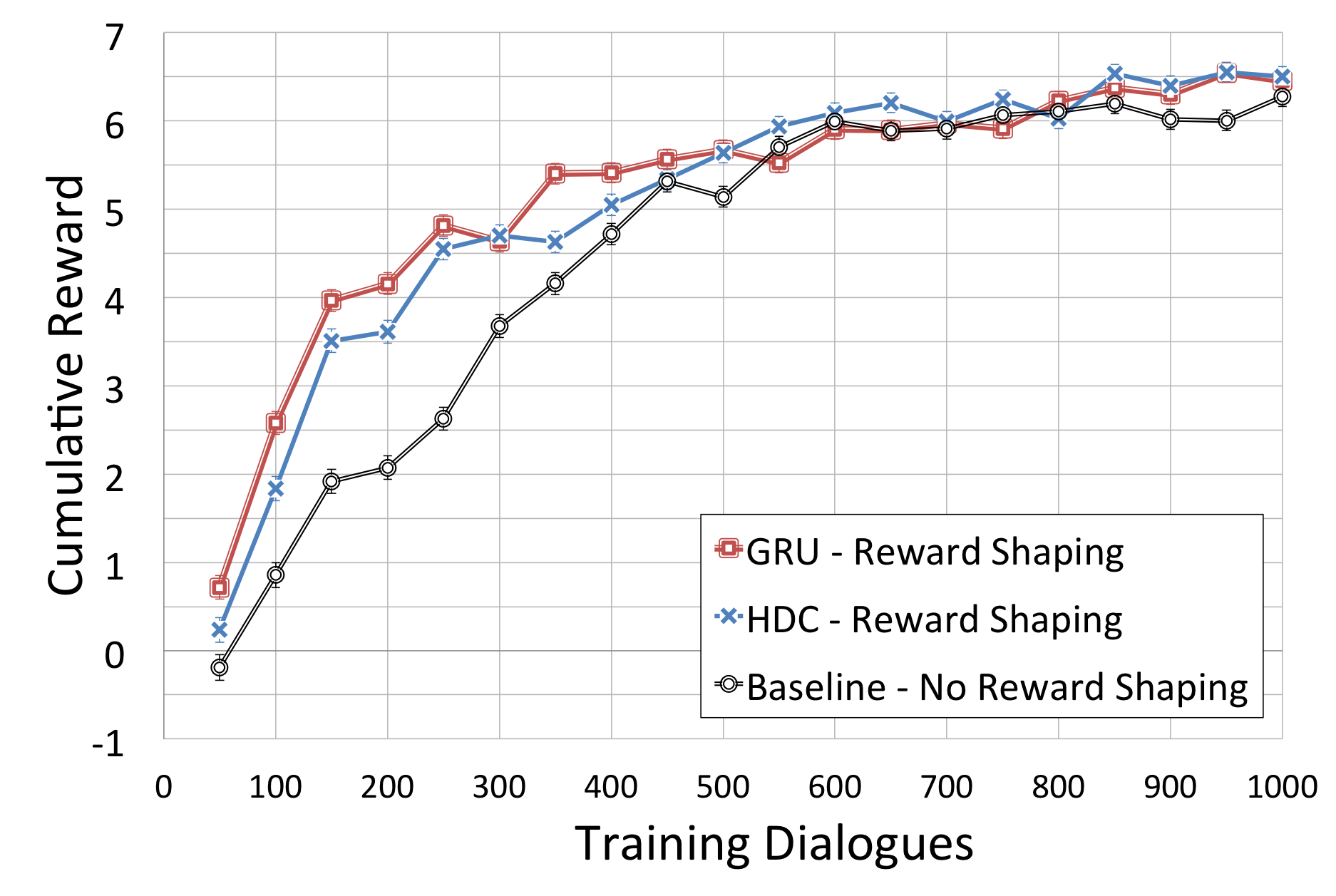}}
\caption{{Policy training via simulated user with (GRU/HDC) and without (baseline) reward shaping. Standard errors are also shown.}} 
\vspace{-3mm}
\label{fig:gp_sim}
\end{figure}

To show the effectiveness of using RNN with GRU for predicting reward shaping potentials, we compare it with the hand-crafted (HDC) method for reward shaping proposed by \newcite{socialRS} that requires prior knowledge of the user's task, and a baseline policy using only the environmental reward. Figure \ref{fig:gp_sim} shows the learning curve of the reward for the three systems. After every 50 training iterations each system was tested with 1000 dialogues and averaged over 10 policies. The simulated user's SER was set to 15\%. 

We see that reward shaping indeed provides the agent with more information, increasing the learning speed. Furthermore, our proposed RNN method further outperforms the hand-crafted system
, whilst also being able to be applied on-line. 


\subsection{Policy Learning with Human Users}
\label{sub:real}

Based on the above results, the same GRU model was selected for training a policy on-line with humans. Two systems were trained with users recruited via Amazon Mechanical Turk: a baseline was trained with only the environmental reward, and another system was trained with an additional shaping reward predicted by the proposed GRU. Learning began from a random policy in all cases. 

\begin{figure}[t]
\centerline{\includegraphics[width=70mm]{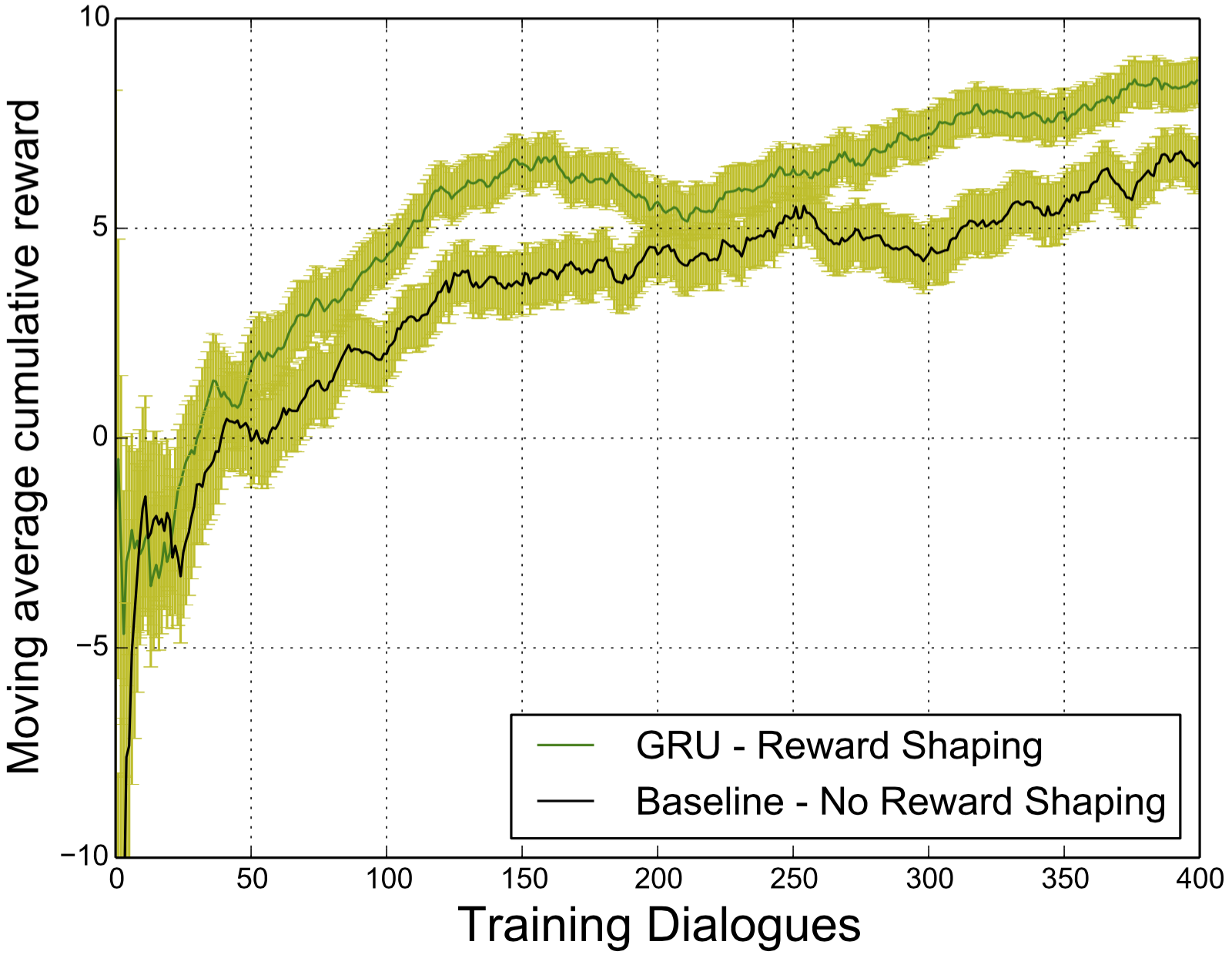}}
\caption{{Learning curves of reward with standard errors during on-line policy optimisation for the baseline (black) and proposed (green) systems.}} 
\vspace{-3mm}
\label{fig:gp_mturk}
\end{figure}

Figure \ref{fig:gp_mturk} shows the on-line learning curve of the reward when training the systems with 400 dialogues. The moving average was calculated using a window of 100 dialogues and each result was averaged over three policies in order to reduce noise. It can be seen that by adding the RNN based shaping reward, the policy learnt quicker in the important initial phase of policy learning.

\section{Conclusions}
\label{sec:conclusions}

This paper has shown that RNN models can be trained to predict the dialogue return with a constraint such that subsequent turn level predictions act as good reward shaping signals that are effective for accelerating policy learning on-line with real users. As in many other applications, we found that gated RNNs such as LSTM and GRU perform a little better than basic RNNs.

In the work described here, the RNNs were trained using a simulated user and this simulator could have been used to bootstrap a policy for use with real users. However our supposition is that RNNs could be trained for reward prediction which are substantially domain independent and hence have wider applications via domain adaptation and extension \cite{MGdistribution2015,polTransRS}. Testing this supposition will be the subject of future work.

 
 

\section{Acknowledgements}
Pei-Hao Su is supported by Cambridge Trust and the Ministry of Education, Taiwan. David Vandyke and Tsung-Hsien Wen are supported by Toshiba Research Europe Ltd, Cambridge Research Lab.

\newpage
\clearpage

\bibliographystyle{acl}
\bibliography{sd2015}

\appendix
\section{Training Constraint/Loss Function}
\label{appendix}
For all RNN models the following MSE loss function is used on a per-dialogue basis:
\begin{equation}
\mbox{MSE}=\left({R}-\sum_{t=1}^{T}r_{t}\right)^2
\end{equation}
where the current dialogue has $T$ turns, ${R}$ is the return and training target, and $r_t$ is the scalar prediction output by the RNN model at each turn.
\end{document}